\documentclass[runningheads]{llncs}
\usepackage{graphicx}
\usepackage{amsmath,amssymb} 
\usepackage{cvpr_eso}
\usepackage{color}
\usepackage{multirow}

\graphicspath{{./Curves/}{./Images/}}

\newcommand{\figref}[1]{Fig.~\ref{#1}}
\newcommand{\tabref}[1]{Tab.~\ref{#1}}

\newcommand{\secref}[1]{Sec.~\ref{#1}}

\begin{document}
\pagestyle{headings}
\mainmatter


\title{MatchBench: An Evaluation of Feature Matchers} 

\author{JiaWang Bian$^1$ Ruihan Yang$^2$ Yun Liu$^2$ Le Zhang$^3$ \\Ming-Ming Cheng$^2$ Ian Reid$^1$ WenHai Wu$^4$}
\authorrunning{Jia-Wang Bian et al.}

\institute{
$^1$University of Adelaide $^2$Nankai University\\ $^3$Advanced Digital Sciences Center $^4$Huawei Technologies\\
}

\maketitle

\begin{abstract}
Feature matching is one of the most fundamental and active research areas in computer vision.
A comprehensive evaluation of feature matchers is necessary, since it would advance both the development of this field and also high-level applications such as Structure-from-Motion or Visual SLAM.
However, to the best of our knowledge, no previous work targets the evaluation of feature matchers while they only focus on evaluating feature detectors and descriptors.
This leads to a critical absence in this field that there is no standard datasets and evaluation metrics to evaluate different feature matchers fairly.
To this end, we present the first uniform feature matching benchmark to facilitate the evaluation of feature matchers.
In the proposed benchmark, matchers are evaluated in different aspects, involving matching ability, correspondence sufficiency, and efficiency.
Also, their performances are investigated in different scenes and in different matching types.
Subsequently, we carry out an extensive evaluation of different state-of-the-art matchers on the benchmark and make in-depth analyses based on the reported results.
This can be used to design practical matching systems in real applications and also advocates the potential future research directions in the filed of feature matching.
\end{abstract}

\section{Introduction}\label{intro}

Feature matching is one of the most fundamental and active research areas in computer vision.
The goal of matching is to build feature correspondences between different views of a scene or object.
The correspondence search provides a basis for image based localization, tracking and reconstruction, so feature matchers are often used in many high-level applications such as Structure-from-Motion~\cite{schonberger2016structure} and Visual SLAM~\cite{davison2007monoslam,mur2015orb}.
Therefore, it is necessary to evaluate feature matchers, which would advance the development of both matching algorithms and related applications. 
However, to the best of our knowledge, \textbf{no previous work targets the evaluation of feature matchers} while they only focus on evaluating feature 
detectors~\cite{mikolajczyk2005comparison,moreels2007evaluation} and 
descriptors~\cite{moreels2007evaluation,mikolajczyk2005performance,heinly2012comparative,hpatches_2017_cvpr,schonberger2017comparative}.
This leads to a critical absence that there is no standard datasets and evaluation metrics to evaluate different feature matchers fairly.

To this end, we propose \textbf{the first uniform feature matching benchmark} to facilitate the analysis of feature matchers.
In the proposed benchmark, matchers are evaluated in three different aspects, involving \emph{matching ability}, \emph{correspondence sufficiency}, and \emph{efficiency}.
Here, the \emph{matching ability} refers to how likely matchers perform correct matching between a pair of images, and the \emph{correspondence sufficiency} refers to how many correspondences matchers proposed when they match an image pair correctly, and the \emph{efficiency} refers to the speed of matching.
They are all critical in high-level applications.
For example, \emph{wrong matchings} or \emph{inadequate correspondences} would cause SfM/SLAM systems to function inappropriately, and \emph{slow matchings} would cause high-level applications to be not able to work at real-time speed.
In order to measure these different aspects, we propose two evaluation metrics, \textbf{SP curves} (with \textbf{AUC score} for showing an overall performance) and \textbf{AP bars}, which respond to the measurement of \emph{matching ability} and \emph{correspondence sufficiency}, respectively.

Instead of reinventing the wheel, our benchmark dataset is constructed by collecting image sequences from existing SfM/SLAM datasets~\cite{Geiger2012CVPR,sturm12iros,strecha2008benchmarking}.
This is because 
\textbf{a)} one goal of this paper is to advance the development of SfM/SLAM~\cite{schonberger2016structure,mur2015orb} by improving the matching techniques, and thus performing the evaluation on SfM/SLAM datasets is the most straightforward;
\textbf{b)} existing SfM/SLAM datasets~\cite{Geiger2012CVPR,sturm12iros,strecha2008benchmarking} are large enough and they cover a wide range of scenes, involving indoor offices, different objects, outdoor street views, and urban buildings.
On the other hand, although images are off-the-shelf, \textbf{we contribute by selecting and re-organizing them} for enabling both \emph{short-baseline} and \emph{wide-baseline} feature matching evaluation, responding to matching problems in Visual SLAM and Structure-from-Motion, respectively.
What's more, \textbf{we make our dataset extensible} by providing easy-to-use tools for re-organizing some popular SLAM/SfM datasets at our format.
It enables researchers to run our evaluation protocol on their own dataset for choosing proper matchers that meet their requirements.

Subsequently, we carry out a comprehensive evaluation of different state-of-the-art feature matchers~\cite{lowe2004distinctive,bay2008speeded,rublee2011orb,liu2012virtual,collins2014analysis,lin2017code,lin2016repmatch,bian2017gms,schonberger2016vote} on the proposed benchmark, and then conduct in-depth analyses based on the results.
This can be used to design practical matching systems in real applications and also advocates the potential future research directions in the filed of feature matching.

The \textbf{contributions} of this paper are as following:
\begin{itemize}
  \item \textbf{a)} we propose \textbf{the first uniform feature matching benchmark} to facilitate the evaluation of feature matchers in different scenes and in different aspects, which enables researchers to develop and evaluate their new algorithms more conveniently.\\
  \item \textbf{b)} we carry out an extensive evaluation of various state-of-the-art matchers and make in-depth analyses, which encourage researchers to design better practical matchers in real applications and also advocate the potential future research directions in the filed of feature matching.
\end{itemize}
On the other hand, the \textbf{novelty} of this paper involves proposing three different aspects for evaluating matchers, designing two evaluation metrics for facilitate the analysis of \emph{matching ability} and \emph{correspondence sufficiency}, and constructing (re-organizing) benchmark datasets for enabling both \emph{short-baseline} and \emph{wide-baseline} feature matching evaluation. 

We organize the paper by giving an overview of feature matchers in~\secref{overview}, introducing evaluation metrics in~\secref{metrics}, constructing benchmark datasets in~\secref{datasets}, and evaluating feature matchers in~\secref{experiments}.
Finally, some discussions of this work are listed in~\secref{discussion}, and conclusions are given in~\secref{conclusion}.

\section{Feature matchers overview}\label{overview}

A typical feature matcher proceeds by \textbf{extracting local features}~\cite{lowe2004distinctive,bay2008speeded,rublee2011orb}, and followed by \textbf{matching features} by using a nearest-neighbor approach~\cite{muja2009fast}, and finally \textbf{selecting good correspondences}~\cite{liu2012virtual,bian2017gms,lin2017code} from the tentative correspondence set.
The selected correspondences would be fed into a RANSAC~\cite{fischler1981random,chum2005matching,torr2000mlesac,rousseeuw2005robust,raguram2013usac} framework for \textbf{fitting a global geometry model}~\cite{nister2004efficient} from them, and the outliers are further rejected by using the estimated model.
Often, the estimated geometry model as well as final correspondences are delivered in Structure-from-Motion~\cite{schonberger2016structure} and Visual SLAM systems~\cite{mur2015orb,davison2007monoslam}.
We give an overview of feature matchers below.

SIFT matcher~\cite{lowe2004distinctive} is a standard method in this field.
It follows the typical pipeline we mentioned above, where FLANN~\cite{muja2009fast} is often used to perform fast (approximated) nearest-neighbor matching, and RATIO~\cite{lowe2004distinctive} is used to select good correspondences which compares the lowest feature distance and the second lowest feature distance for recognizing good ones.
SIFT matcher is widely used in different applications, and we \textbf{regard it to be the baseline} for analyzing other matchers.

There are two main research directions for boosting matchers' performances or efficiencies, including designing \textbf{better local features}~\cite{bay2008speeded,rublee2011orb,alcantarilla2012kaze,alcantarilla2011fast,leutenegger2011brisk,alahi2012freak,simonyan2014learning} and \textbf{better matching solutions}~\cite{liu2012virtual,lin2016repmatch,lin2017code,collins2014analysis,bian2017gms}.
Here, \emph{local features} are reviewed and evaluated in many previous works~\cite{mikolajczyk2005performance,moreels2007evaluation,heinly2012comparative,hpatches_2017_cvpr,schonberger2017comparative}, but \emph{matching solutions} are few discussed.
Therefore, we focus on introducing the latter below.

\textbf{Graph matchers}~\cite{leordeanu2005spectral,zhou2013deformable,zhou2012factorized,collins2014analysis} search for geometry consistent correspondences between two sets of features, rather than performing nearest-neighbor matching and selecting good correspondences like a typical matching system.
They optimize a global consistency score and can cope with higher-order constraints (involving more than one match).
However, they are not well suited for a high outlier rate, and their time and space complexity grows exponentially with the order, which limits in real applications to a few hundred feature points.

\textbf{KVLD matcher}~\cite{liu2012virtual} proposes a virtual Line descriptor and a semi-local matching method based on this descriptor for correspondence selection.
It makes good use of constraints in both photometry and geometry, and correspondences that pass the verification in both domains are recognized to be good.
The methods works well in strong-texture scenes while suffers in weak-texture scenes because in this scenario photometry-based solutions may function inappropriately.

\textbf{CODE matcher}~\cite{lin2017code} proposes an optimization based approach for finding a globally smooth correspondence set.
Employing powerful ASIFT~\cite{morel2009asift} feature, it performs ultra-robust wide-baseline matching and proposes sufficient correspondences.
Based on CODE, \textbf{RepMatch matcher}~\cite{lin2016repmatch} proposes a geometry-aware approach to tackle the challenges of repetitive structures.
It improves the performance again but introduces a higher complexity at the same time.
However, these two matchers~\cite{lin2016repmatch,lin2017code} have huge computational costs, although they are very powerful.

\textbf{GMS matcher}~\cite{bian2017gms} proposes a correspondence selection method called \emph{grid-based motion statistics}.
It is fast and robust to recognize good correspondences.
Adopting cheap and rich ORB~\cite{rublee2011orb} feature, the whole matcher can perform high-quality matching while achieving real-time performances.

Finally, with respect to \emph{the number of correspondences}, matchers can also be divided into two classes, \textbf{sparse feature matchers}, and \textbf{rich feature matchers}.
Here, CODE~\cite{lin2017code}, RepMatch~\cite{lin2016repmatch}, and GMS~\cite{bian2017gms} fall into the group of \emph{rich feature matchers}, since their output correspondences are much more than other \emph{sparse matchers}.
They are all recently proposed and show high-quality feature matching.
We compare these two classes of matchers in our evaluation.

\section{Evaluation metrics}\label{metrics}
The inputs of a matcher are two images and the outputs are correspondences between them.
It sounds straightforward to \textbf{benchmark the output correspondences}, but one may \textbf{find that it is impractical} due to the difficulty of generating the highly-quality (accurate and dense) ground truth.
To our knowledge, there are two methods for ground-truth correspondences generation:
\textbf{a)}
The first approach is projecting a pixel in one image to other images by using Homography (see details in~\cite{hartley2003multiple}).
However, this can only be used in a planar scene \textbf{but not be applicable in a generic non-planar scene}.
\textbf{b)}
The other approach enables projection in non-planar scenes by using internal camera parameters (calibration matrix), external camera parameters (camera poses), and depth images.
However, the method turns out to be \textbf{lacking of density and precision} due to the low-quality (sparse and low-precision) depth, leading to less conclusive results.

To this end, we propose to feed correspondences into a pose estimator and benchmark the results of pose estimation, instead of directly evaluating correspondences.
In this design, the \textbf{pose error} (compared with the ground-truth pose) implies how well an image pair is matched, and a matched pair would \textbf{be regard to be correct} if its pose error is less than certain threshold.
For estimating the relative camera pose $T$, we firstly estimate the essential matrix $E$ from correspondences $C$ and internal camera parameters (calibration matrix) $K$:
\begin{equation}
E \leftarrow {C,K}
\end{equation}
Alternatively, we can also estimate the fundamental matrix $F$ from correspondences $C$ and convert it to $E$ given $K$:
\begin{equation}
F \leftarrow C
\end{equation}
\begin{equation}
E = K^{-1} \cdot F \cdot K
\end{equation}
Then we get:
\begin{equation}
T \leftarrow \{C, E, K\}
\end{equation}
More details about pose estimation can be referenced in~\cite{nister2004efficient,hartley2003multiple}.

In order to \textbf{measure the correctness of a matched pair}, we spit $T$ into a \emph{rotation matrix} $R$ and a \emph{translation vector} $t$, and then compare them with the ground-truth $R_{gt}$ and $t_{gt}$.
This leads to a \emph{rotational error} $e_r$ and \emph{translational error} $e_t$.
Here, both errors are in degrees.
Specifically, $e_r$ is computed from the transformation matrix from $R_{gt}$ to $R$ as did in KITTI~\cite{Geiger2012CVPR}, and $e_t$ is the angle between vectors $t_{gt}$ and $t$.
Note that two translational vectors are in different scales because the scale cannot be estimated given monocular image pairs (see details in~\cite{hartley2003multiple}).
We set the \emph{camera pose error} to be:
\begin{equation}
e = max (e_r , e_t)
\end{equation}
Then an image pair would \textbf{be recognized as a correct match} if its pose error $e$ is less than certain threshold. 

Given the above method to verify a matched pair, we further propose two metrics for benchmarking a matcher: \textbf{SP} (Success ratio / Pose error threshold) curves and \textbf{AP} (Averaged number of correspondences / Pose error threshold) bars.
\textbf{SP} curves show the change of \emph{success ratio}, the percentage of correctly matched pairs to all image pairs, with increasing pose error thresholds.
This responds to \emph{matching ability} measurement.
\textbf{AP} bars illustrate the mean number of correspondences averaged over correctly matched pairs (the threshold of $5$ degrees is used here).
It measures the \emph{correspondence sufficiency} of matchers.
Besides, for showing an overall \emph{matching ability} of matchers, we propose to compute the \textbf{AUC score}(Area Under Curve) of \textbf{SP curves}.
As the pose error thresholds are discrete, the \emph{AUC score} of matchers equals to the mean value of its success ratios on all pose error thresholds.

\section{Benchmark dataset}\label{datasets}

One goal of this paper is to evaluate feature matchers for advancing the development of Visual SLAM/SfM~\cite{schonberger2016structure,davison2007monoslam,mur2015orb}. 
Therefore, rather than reinventing the wheel, we construct our benchmark dataset by collecting image sequences from existing SLAM/SfM datasets, including TUM SLAM dataset~\cite{sturm12iros}, KITTI odometry benchmark~\cite{Geiger2012CVPR}, and Strecha SfM dataset~\cite{strecha2008benchmarking}
They not only provide real-world image sequences of different scenes, but also provide the precise camera trajectory (camera positions) as the ground truth.
Besides, we split the dataset into two portions, involving \emph{short-baseline matching} and \emph{wide-baseline matching} portions.
In different portions, the methods to construct image pairs are different.
We introduce the dataset below.

\begin{figure}[ht]
\begin{center}
      \includegraphics[width=0.8\linewidth]{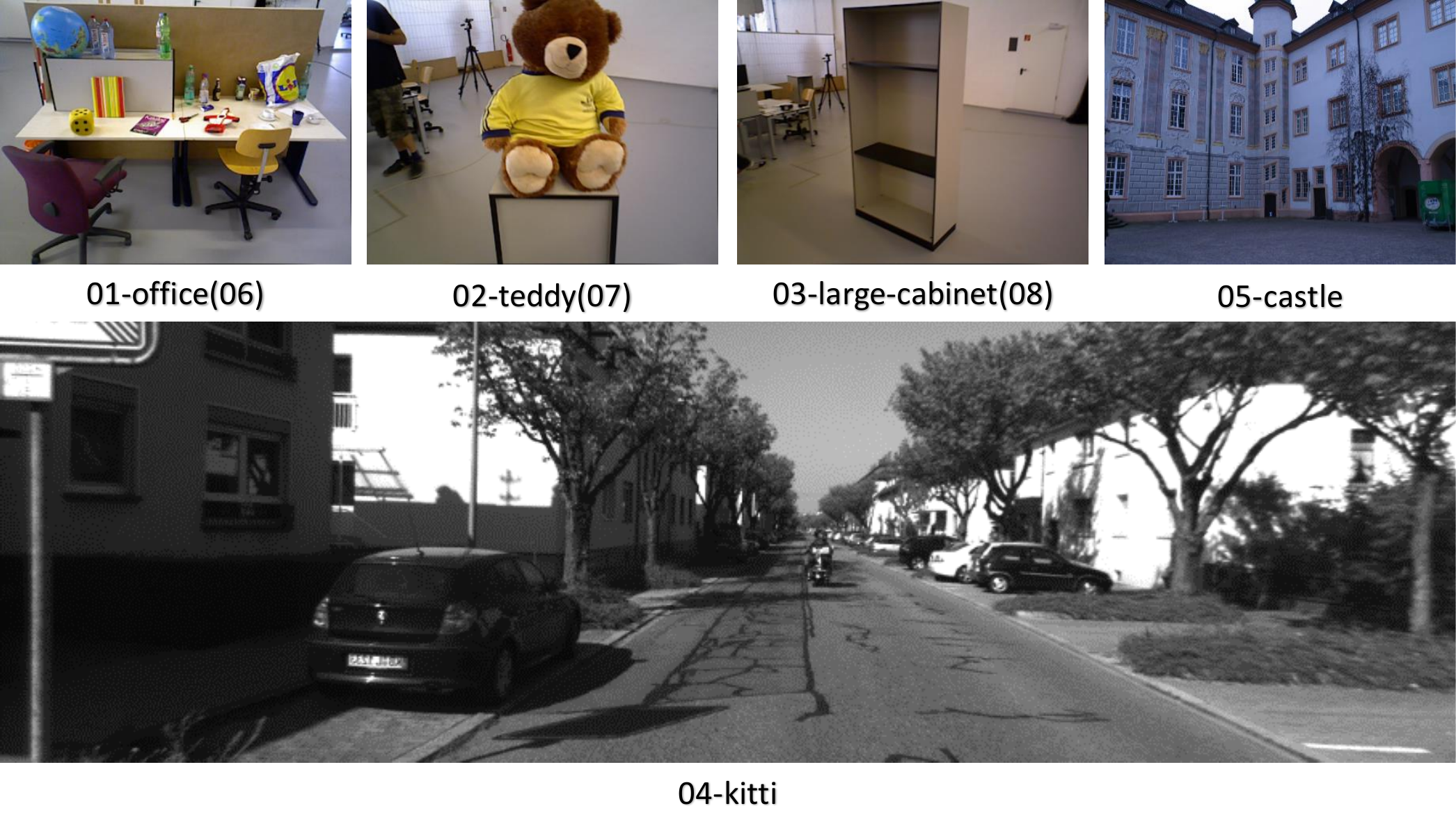}
\end{center}
   \caption{The screen-shots of the selected images sequences. The sequences 01-03 (06-08), 04, 05 are collected from TUM dataset~\cite{sturm12iros}, KITTI dataset~\cite{Geiger2012CVPR}, and Strecha dataset~\cite{strecha2008benchmarking}, respectively.
   }
\label{fig-dataset}
\end{figure}

\textbf{Dataset description}.
Our dataset contains eight image sequences with the first four sequences for \emph{short-baseline matching} evaluation and the last four sequences for \emph{wide-baseline matching} evaluation.
They are selected from three datasets, 
including TUM dataset~\cite{sturm12iros} where videos of indoor scenes are captured at $30fps$ and the sensor resolution is $640 \times 480$, 
KITTI dataset~\cite{Geiger2012CVPR} where video sequences of street views are captured at $10fps$ and the image resolution is $376\times1241$,
and Strecha dataset~\cite{strecha2008benchmarking} where authors provide image sequences of urban buildings and the resolution is $3072\times2048$.
The screen-shot and description of selected image sequences are illustrated in \figref{fig-dataset} and \tabref{tab-dataset}, respectively.
Here, sequence 04 is from KITTI dataset~\cite{Geiger2012CVPR} and sequence 05 is from Strecha dataset~\cite{strecha2008benchmarking}.
They are easier than other sequences, since others are from TUM dataset~\cite{sturm12iros} where the texture of scenes is weaker and the image resolution is lower. 
Especially, sequence 02(07) and sequence 03(08) are challenging, as the former captures a non-planar object and the latter captures a low-texture object.

\begin{table}[ht]
\centering
\caption{The description of the selected image sequences.}
\label{tab-dataset}
\resizebox{0.96\textwidth}{!}{
\begin{tabular}{ c|lcccc }
   \hline
   Type & Sequences     & Images  & Pairs     & Resolution & Property   \\
   \hline
   \multirow{ 4}{*}{\rotatebox{90}{Short}} & 01-office   & 2583   & 2310 & $480 \times 640$ &indoor, office\\
   &02-teddy   & 2405  & 2234       & $480 \times 640$ &indoor, non-planar\\
   &03-large-cabinet    & 1006  & 938    & $480 \times 640$ & indoor, weak texture\\
   &04-kitti    & 4542  & 3632       & $376\times1241$ & outdoor, street-view\\  
   \hline
   \multirow{ 4}{*}{\rotatebox{90}{Wide}} & 05-castle   & 30   & 435       & $3072\times2048$ & outdoor, urban \\ 
   &06-office-wide   &173  & 1512       & $480 \times 640$ & same with 01 \\ 
   &07-teddy-wide    & 161   &1404     & $480 \times 640$ & same with 02 \\
   &08-large-cabinet-wide  & 68   & 567    & $480 \times 640$ & same with 03 \\
   \hline
\end{tabular}%
}
\vspace{-0.1in}
\end{table}

\textbf{Short-baseline matching portion}.
Three sequences (01-03) are from TUM dataset~\cite{sturm12iros} and one sequence (04) is from KITTI dataset~\cite{Geiger2012CVPR}.
Every video sequence is divided into $m$ non-overlapping fragments and each fragment contains $k$ frames.
In each fragment, the first frame is set to be the reference image and other $k-1$ frames will be matched to it.
Here, $k$ is set to be $15$ for sequences 01-03 and be $5$ for sequence 04, since they are captured at $30$ fps and $10$ fps, respectively.
It means the time length of each fragment is $0.5$ seconds.
This results in $n-m$ image pairs in each sequence, where $n$ is the number of images in the sequence.

\textbf{Wide-baseline matching portion}.
The fifth sequence (05-castle) is selected from Strecha~\cite{strecha2008benchmarking} dataset, and other sequences 06-08 are sub-sampled from sequences 01-03.
For the sequence 05, we run all possible $n*(n-1)/2$ pairs, where $n$ is the number of images in the sequence.
For sequences 01-03, we extract the first image of every fragment (where each fragment contains $15$ frames) in each sequence, leading to sequences 06-08.
Then for each sequence, every image is matched to the next at most $9$ images.
Since the the sensor is $30fps$, every image is matched to the next frames captured within 5 seconds.
This is based on our observation that most pairs beyond 5 seconds are with no overlap.
Note that in this portion not all pairs are with overlap, but this doesn't influence the relative performance of different matchers because false pairs are nearly impossible to be "matched and estimated correctly" by any matcher.

\section{Experiments}\label{experiments}

We perform exhaustive evaluation of different feature matchers in this section.
As described in~\secref{intro}, matchers are evaluated in terms of \emph{matching ability}, \emph{correspondence sufficiency}, and \emph{efficiency}.
They are also evaluated in different type of matching tasks, involving \emph{short-baseline matching} and \emph{wide-baseline matching}.
Evaluation settings, experimental results, and analyses are given in the following sections.

\subsection{Evaluation setting}

\textbf{Evaluated matchers}.
For a comprehensive evaluation, we collect various state-of-the-art matchers.
They fall into two main categories, \emph{distinctive local features} and \emph{powerful matching solutions}, as described in~\secref{overview}.
\textbf{The first category} includes SIFT~\cite{lowe2004distinctive}, SURF~\cite{bay2008speeded}, ORB~\cite{rublee2011orb}, BRISK~\cite{leutenegger2011brisk}, KAZE~\cite{alcantarilla2012kaze}, AKAZE~\cite{alcantarilla2011fast}, DLCO~\cite{simonyan2014learning}, FREAK~\cite{alahi2012freak}, BinBoost~\cite{trzcinski2013boosting}, LATCH~\cite{levi2016latch}, and DAISY~\cite{tola2010daisy} total \textbf{$11$} methods.
Here, the last five methods (DLCO, FREAK, BinBoost, LATCH, DAISY) only provide feature descriptors and no detector is available.
Therefore, we concatenate them (except for FREAK~\cite{alahi2012freak}) with the SIFT~\cite{lowe2004distinctive} detector, and combine FREAK descriptor with  SURF~\cite{bay2008speeded} detector as suggested in OpenCV samples.
These features follow the classical matching pipeline that features are matched by using a nearest-neighbor approach and correspondences are selected by using RATIO (the threshold is 0.8 as widely used in applications).
\textbf{The second category} includes KVLD~\cite{liu2012virtual}, GAIM~\cite{collins2014analysis}, CODE~\cite{lin2017code}, RepMatch~\cite{lin2016repmatch}, and GMS~\cite{bian2017gms} matchers.
A brief description of different matchers can be seen in \secref{overview}.

The problem associated \emph{short-baseline matching} involves video-based applications such as Visual SLAM~\cite{davison2007monoslam,mur2015orb} where the efficiency is quite critical.
It would be less meaningful to evaluate a slow matcher if it could not be integrated into real-time applications.
Therefore, we exclude slow matchers (KVLD, GAIM, CODE, and RepMatch) in \emph{short-baseline matching} portion, as they seem far away from enabling fast matching even though GPU is available.
For \emph{wide-baseline matching}, all matchers are evaluated.

\textbf{Camera pose estimation}.
We adopt two pose estimators for camera pose estimation.
The first one is from OpenCV library which implements five-points~\cite{nister2004efficient} method for essential matrix estimation within a robust RANSAC~\cite{fischler1981random} framework.
The estimator is well-tuned and widely used for estimating relative camera pose from a set of correspondences.
However, we empirically find that this estimator doesn't work well for rich feature matchers (CODE~\cite{lin2017code}, RepMatch~\cite{lin2016repmatch}, GMS~\cite{bian2017gms}), as their output correspondences are much more than traditional sparse matchers.
Therefore, we propose to use the pose estimator built in RepMatch~\cite{lin2016repmatch} for these three rich matchers.
We also try to use this estimator with other sparse matchers, like SIFT~\cite{lowe2004distinctive}.
However, the results show that the OpenCV estimator is consistently better.
Therefore, for sparse matchers, we still use the OpenCV estimator.

\textbf{Implementation details}.
The implementation of all local features is from OpenCV library.
We use their default parameters for extracting features, except for ORB~\cite{rublee2011orb} feature.
Here, the default \emph{nfeatures} of ORB implementation is $500$ which limits the maximum number of detected features.
We manually assign a big number ($100,000$) to it for breaking this limitation.
Note that the number of detected features are often much more lower than this value in practice.
On the other hand, in order to match features, we adopt FLANN matcher~\cite{muja2009fast} with Euclidean distance for real-value features (SIFT, SURF, KAZE, DLCO, DAISY) and adopt Brute-force matcher with Hamming distance for matching binary features (ORB, BRISK, AKAZE, FREAK, BinBoost, LATCH) for the best trade-off between performances and efficiency.
This is a widely used setting in feature matching.
What's more, for others matchers, we follow the default setting provided by authors.
Specifically, KVLD~\cite{liu2012virtual} adopts SIFT feature; 
CODE~\cite{lin2017code} and RepMatch~\cite{lin2016repmatch} employ ASIFT~\cite{morel2009asift} feature;
GAIM~\cite{collins2014analysis} simulates images and extracts SURF~\cite{bay2008speeded} feature;
GMS~\cite{bian2017gms} adopts ORB~\cite{rublee2011orb} feature (extracting at most $10K$ interest points).

\textbf{Speed testing}.
For comparing the time consumption of different matchers fairly, we run all algorithms in one computer where CPU is Intel i7-6700K and GPU is NVIDIA GTX 1080.
The first $200$ image pairs in sequence 01 are used to evaluate the speed of matchers, and the averaged time consumption of matchers is reported in \tabref{tab-speed}.
Note that most feature detection and nearest-neighbor matching methods can be accelerated by GPU while correspondence selection approaches are not trivial to be accelerated.
Therefore, the latter may be the bottleneck in real applications.

\subsection{Evaluation results and analyses}

\begin{figure}[!ht]
\centering
\begin{tabular}{ cc }
	\includegraphics[width=0.48\linewidth]{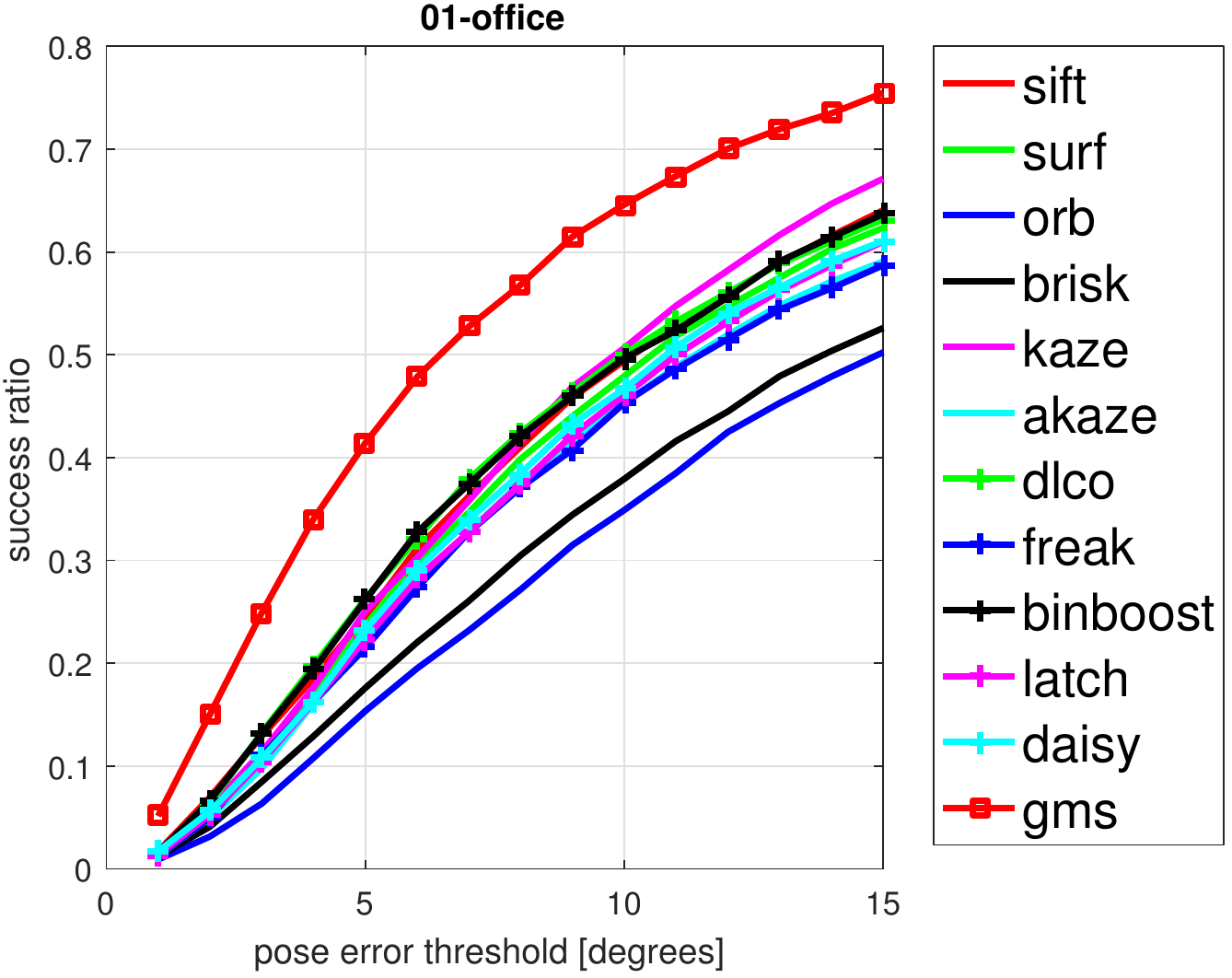} &
    \includegraphics[width=0.48\linewidth]{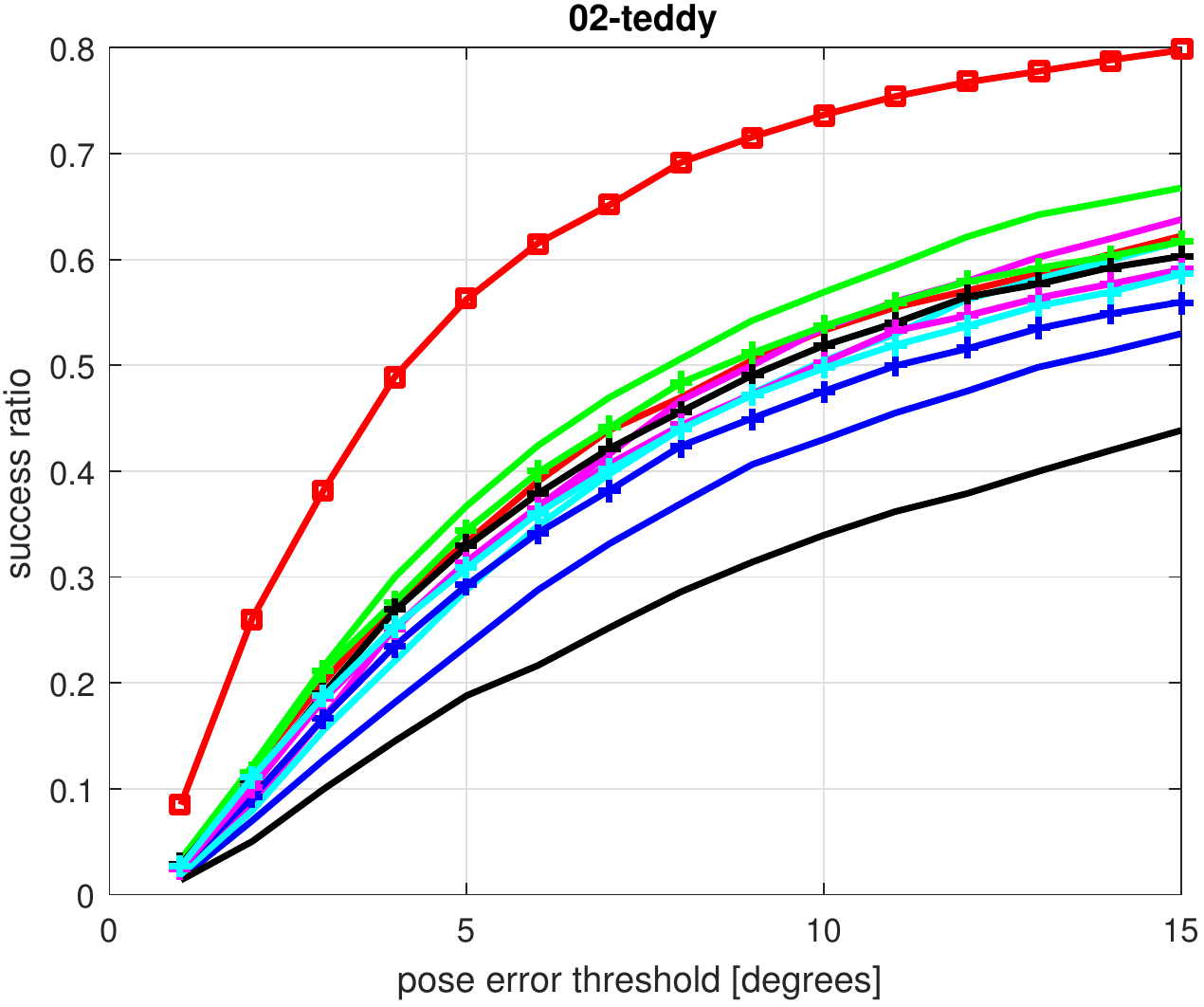} \\
    (a)  & (b) \\
	\includegraphics[width=0.48\linewidth]{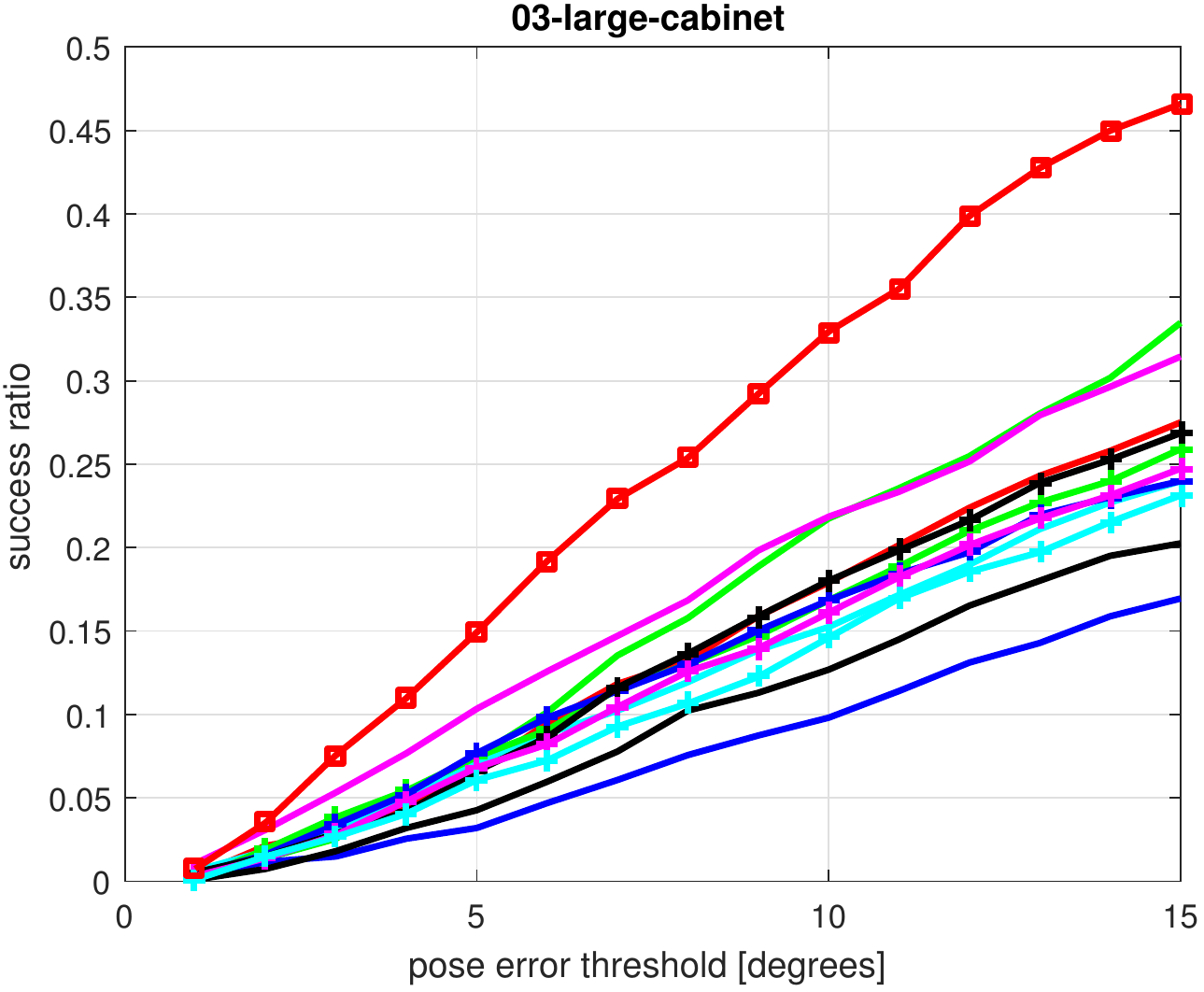} &
    \includegraphics[width=0.48\linewidth]{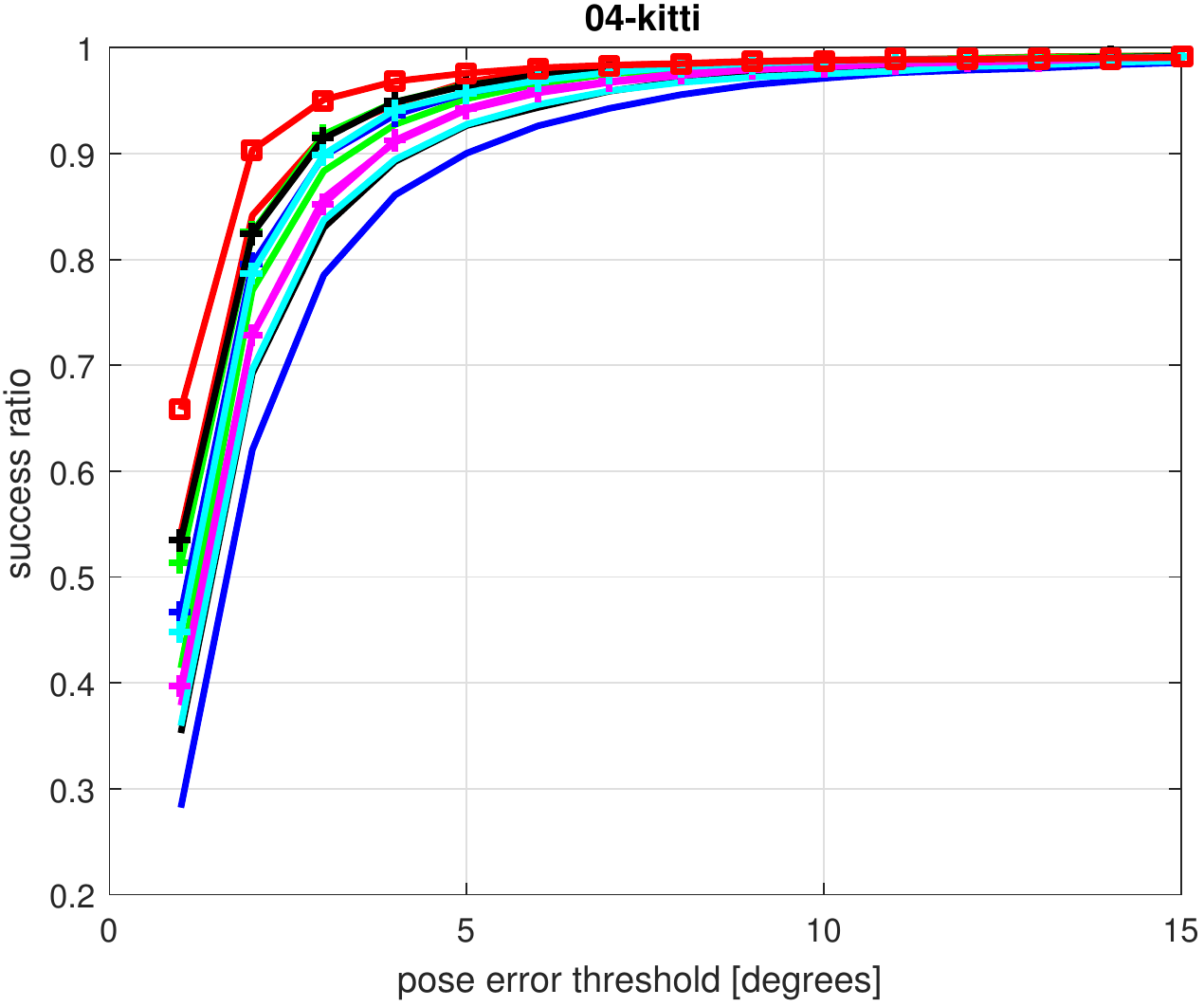} \\
    (c)  & (d)  \\
    \multicolumn{2}{c}{\includegraphics[width=0.98\linewidth]{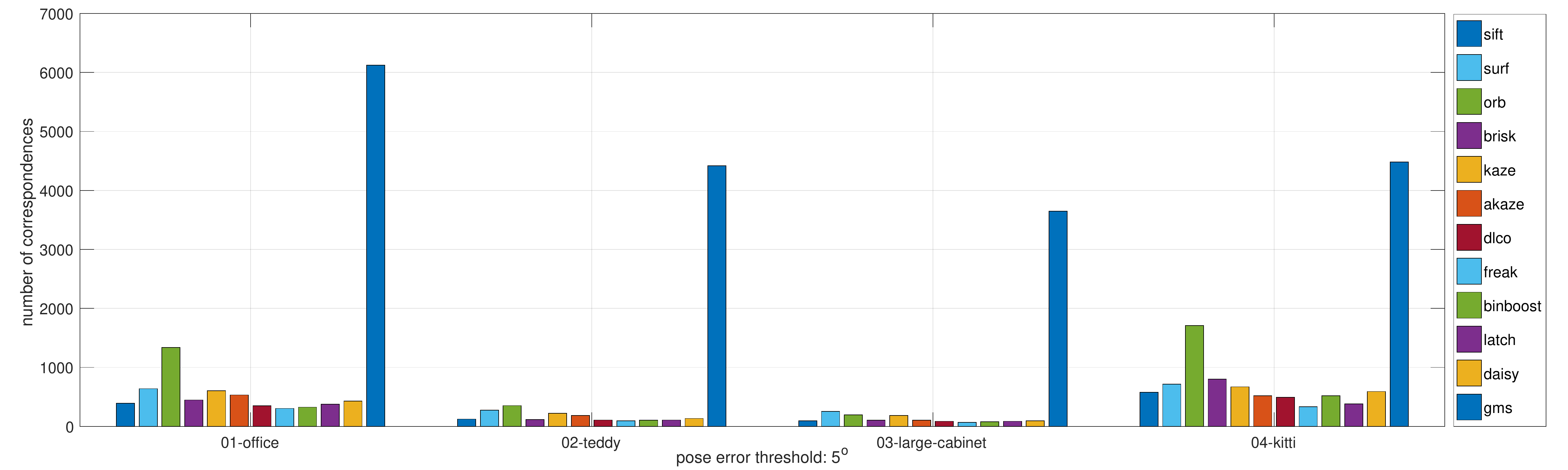}} \\
    \multicolumn{2}{c}{(e)}
\end{tabular}
\caption{
Evaluation results on the short-baseline matching portion. 
\emph{SP} curves show the \emph{success ratio} of matchers with changing \emph{pose error thresholds} and \emph{AP} bars show the \emph{number of correspondences} which is averaged over correctly matched pairs.
}
\label{fig-results1}
\end{figure}

\begin{figure}[!ht]
\centering
\begin{tabular}{ cc }
	\includegraphics[width=0.48\linewidth]{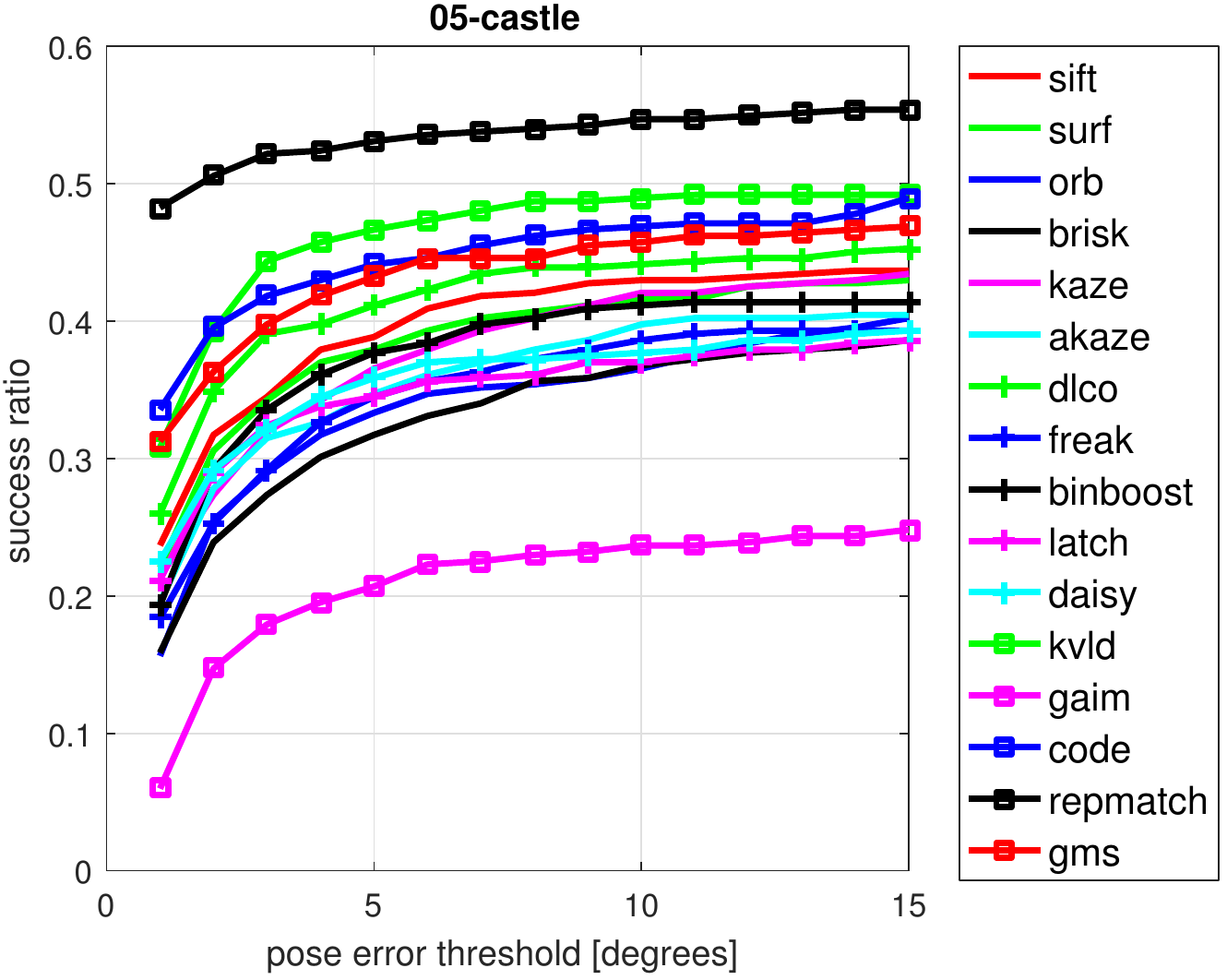} &
    \includegraphics[width=0.48\linewidth]{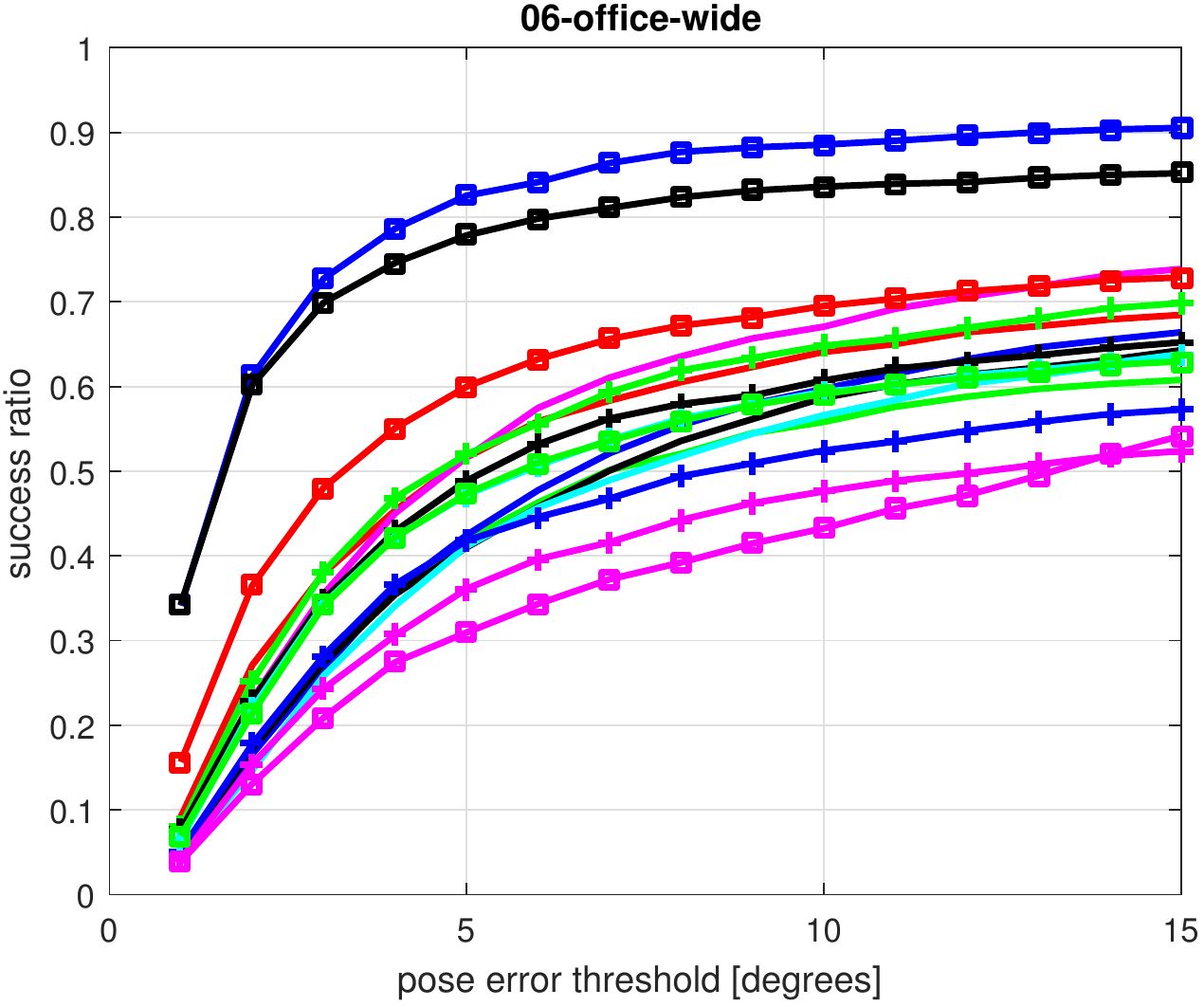} \\
    (a) & (b) \\

	\includegraphics[width=0.48\linewidth]{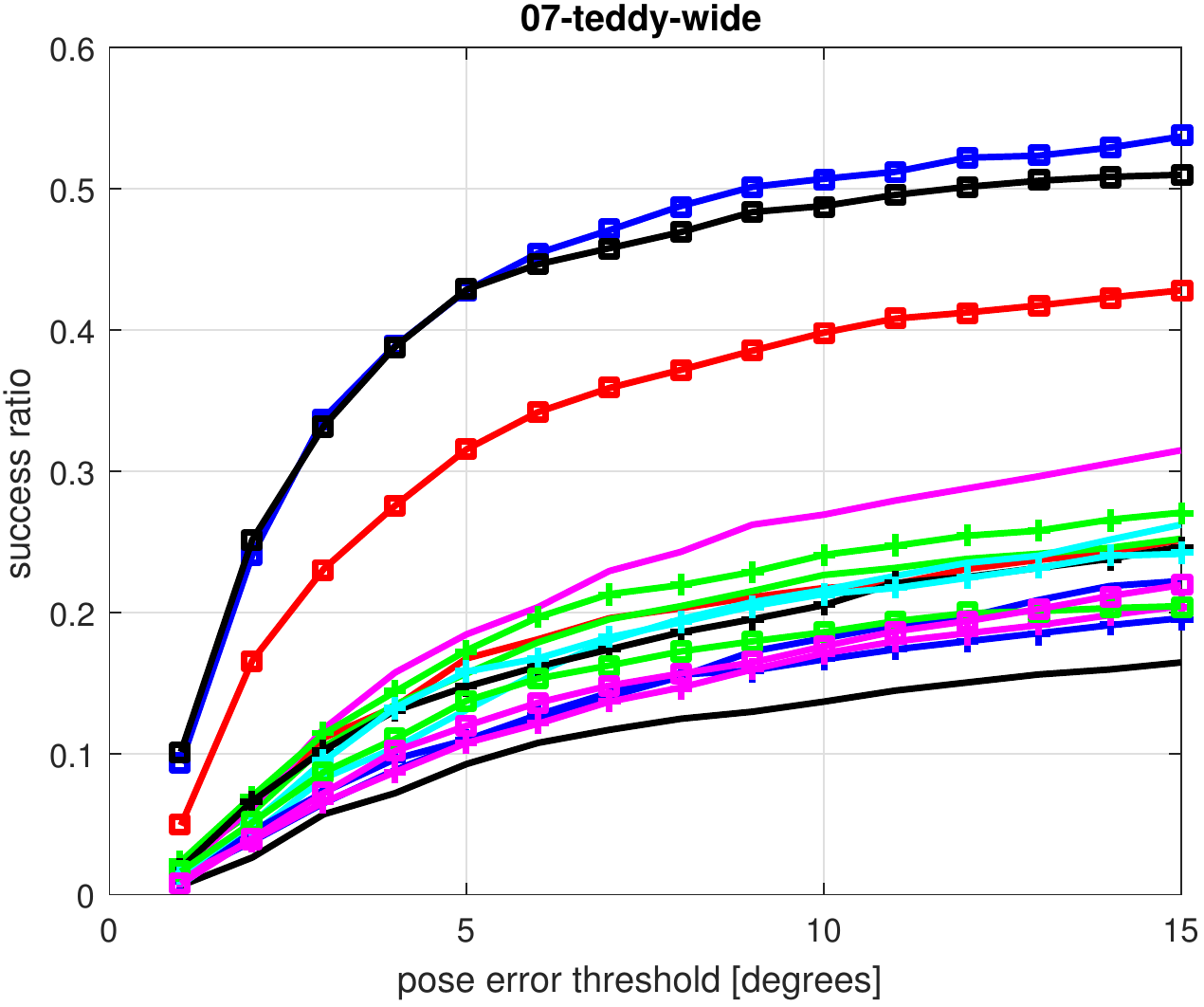} &
    \includegraphics[width=0.48\linewidth]{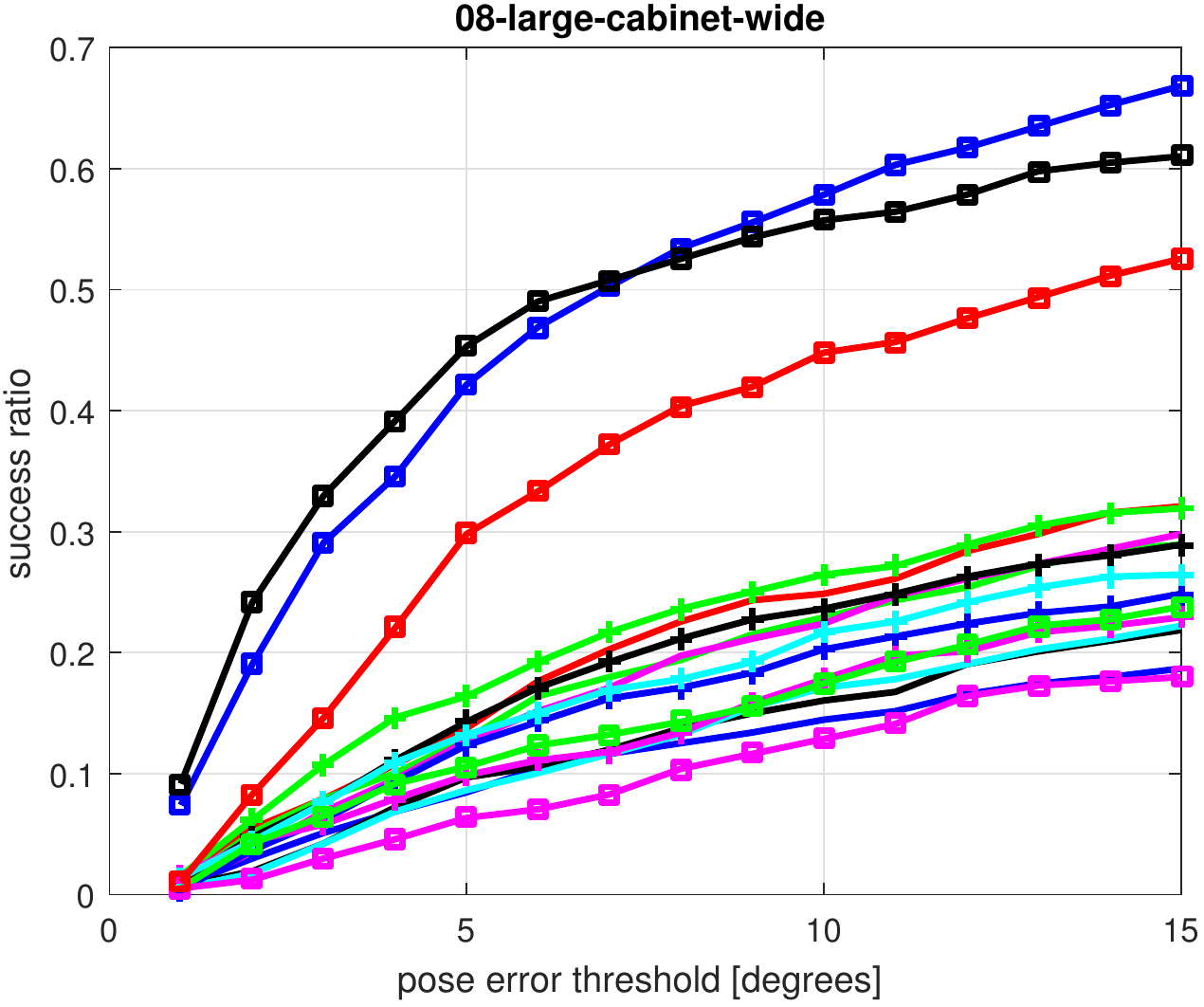} \\
    (c) & (d) \\
    \multicolumn{2}{c}{\includegraphics[width=0.98\linewidth]{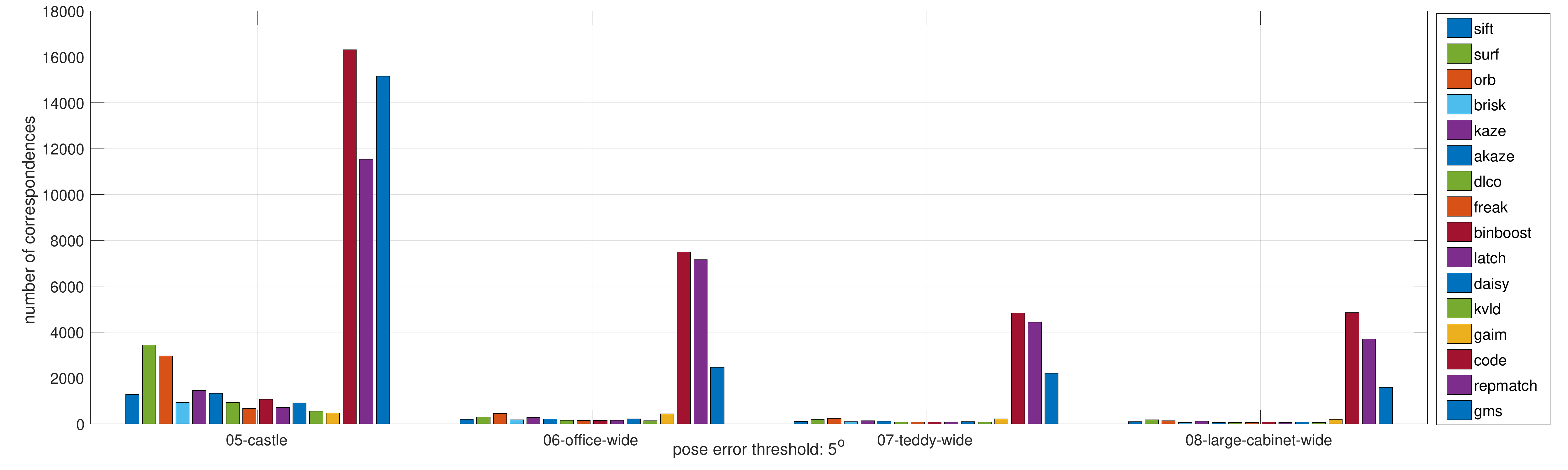}} \\
    \multicolumn{2}{c}{(e)}
\end{tabular}
\caption{
Evaluation results on the wide-baseline matching portion.
\emph{SP} curves show the \emph{success ratio} of matchers with changing \emph{pose error thresholds} and \emph{AP} bars show the \emph{number of correspondences} which is averaged over correctly matched pairs.
}
\label{fig-results2}
\end{figure}

\begin{table}
\centering
\caption{The AUC score of matchers. The best three methods in each sequence are labeled in red, green, and blue colors, respectively. Besides, matchers which outperform the baseline (SIFT matcher) are labeled in bold font.}
\small
\begin{tabular}{|c|c|c|c|c|c|c|c|c|}
\hline
\multirow{2}{*}{Matchers} & \multicolumn{4}{c|}{short-baseline portion} &\multicolumn{4}{c|}{wide-baseline portion}\\
\cline{2-9}
 & 01 & 02 & 03 & 04 & 05 & 06 & 07 & 08 \\
\hline
\textbf{SIFT}  & 0.375& 0.415& 0.137& \textcolor{green}{0.939}& 0.396& 0.538& 0.179& 0.198\\
\hline
SURF  & 0.360& \textbf{\textcolor{green}{0.449}}& \textbf{\textcolor{blue}{0.159}}& 0.919& 0.385& 0.456& \textbf{0.180}& 0.180\\
\hline
ORB  & 0.265& 0.328& 0.078& 0.874& 0.338& 0.480& 0.141& 0.115\\
\hline
BRISK  & 0.288& 0.260& 0.098& 0.897& 0.329& 0.468& 0.110& 0.127\\
\hline
KAZE  & \textbf{\textcolor{green}{0.382}}& 0.408& \textbf{\textcolor{green}{0.167}}& 0.908& 0.377& \textbf{0.557}& \textbf{0.215}& 0.177\\
\hline
AKAZE  & 0.340& 0.388& 0.121& 0.897& 0.358& 0.456& 0.169& 0.127\\
\hline
DLCO  & \textbf{\textcolor{blue}{0.379}}& \textbf{\textcolor{blue}{0.420}}& 0.131& 0.935& \textbf{0.415}& \textbf{0.543}& \textbf{0.195}& \textbf{0.210}\\
\hline
FREAK  & 0.339& 0.370& 0.128& 0.927& 0.348& 0.435& 0.134& 0.156\\
\hline
BinBoost  & \textbf{0.378}& 0.404& 0.134& \textcolor{blue}{0.936}& 0.375& 0.508& 0.170& 0.185\\
\hline
LATCH  & 0.348& 0.392& 0.124& 0.909& 0.349& 0.389& 0.134& 0.138\\
\hline
DAISY  & 0.354& 0.388& 0.112& 0.925& 0.356& 0.493& 0.171& 0.169\\
\hline
KVLD  & / & / & / & / & \textbf{\textcolor{green}{0.463}}& 0.492& 0.150& 0.142\\
\hline
GAIM  & / & / & / & / & 0.210& 0.360& 0.142& 0.100\\
\hline
CODE  & / & / & / & / & \textbf{\textcolor{blue}{0.447}}& \textbf{\textcolor{red}{0.809}}& \textbf{\textcolor{red}{0.436}}& \textbf{\textcolor{red}{0.476}}\\
\hline
RepMatch  & / & / & / & / & \textbf{\textcolor{red}{0.535}}& \textbf{\textcolor{green}{0.766}}& \textbf{\textcolor{green}{0.425}}& \textbf{\textcolor{green}{0.473}}\\
\hline
GMS  & \textbf{\textcolor{red}{0.508}}& \textbf{\textcolor{red}{0.605}}& \textbf{\textcolor{red}{0.251}}& \textbf{\textcolor{red}{0.955}}& \textbf{0.433}& \textbf{\textcolor{blue}{0.605}}& \textbf{\textcolor{blue}{0.332}}& \textbf{\textcolor{blue}{0.347}}\\
\hline
\end{tabular}
\label{tab-auc}
\end{table}

The experimental results of \emph{short-baseline matching} and \emph{wide-baseline matching} are illustrated in \figref{fig-results1} and \figref{fig-results2}, respectively. 
The \emph{AUC score} of matchers is shown in \tabref{tab-auc}, and the \emph{time consumption} of matchers is shown in \tabref{tab-speed}.
These results enable us to analyze the \emph{matching ability}, \emph{correspondence sufficiency}, as well as \emph{efficiency} of different matchers.
We make the following analyses.

\textbf{a) the experimental data (image size, scene type, and etc) influences the performance of matchers significantly.} 
Seeing \figref{fig-results1} or \tabref{tab-auc}, one can find that the \emph{matching abilities} of matchers are high in sequence 04, and they are significantly lower in sequences 01-03.
At the same time, the performance gap of different matchers is narrow in sequence 04 and it is wide in sequences 01-03.
Due to this, we may regard \emph{matching ability} to be the most vital factor to choose good matchers in sequences 01-03, but we may pay more attention on the \emph{efficiency} or \emph{correspondence sufficiency} of matchers in sequence 04 for the best trade-off.
Therefore, we suggest researchers \textbf{re-organizing their own dataset and running our evaluation protocol on it for selecting appropriate matchers} before developing an real application.

\textbf{b) rich feature matchers vs sparse feature matchers}.
Three matchers (CODE~\cite{lin2017code}, RepMatch~\cite{lin2016repmatch}, and GMS~\cite{bian2017gms}) fall into the first class while other matchers fall into the second class.
First, in terms of \emph{matching ability}, rich matchers outperform the sparse matchers.
This is demonstrated in \tabref{tab-auc} where GMS matcher~\cite{lin2017code} outperforms other (sparse) matchers consistently in \emph{short-baseline portion} and rich matchers (CODE~\cite{lin2017code}, RepMatch~\cite{lin2016repmatch}, and GMS~\cite{bian2017gms}) outperform others in \emph{wide-baseline portion} (except for the case that KVLD matcher~\cite{liu2012virtual} slightly outperforms CODE and GMS in sequence 05).
Second, with respect to \emph{correspondence sufficiency} (see \figref{fig-results1} or \figref{fig-results2}), rich matchers naturally outperform sparse matchers.
Third, with regard to \emph{efficiency}, CODE~\cite{lin2017code} and RepMatch~\cite{lin2016repmatch} are much more slower than most sparse matchers (except for GAIM~\cite{collins2014analysis}) even though GPU is adopted, but GMS~\cite{bian2017gms} can show real-time performances.

\textbf{c) local feature extractors.}
We regard SIFT feature~\cite{lowe2004distinctive} to be the baseline for analyzing other local features.
First, with regard to \emph{matching ability} (see \tabref{tab-auc}), three features (SURF~\cite{bay2008speeded}, KAZE~\cite{alcantarilla2012kaze}, DLCO~\cite{simonyan2014learning}) show equivalent and higher performances than SIFT feature, while other features are not as good as that.
Second, in terms of \emph{correspondence sufficiency} (see \figref{fig-results1} or \figref{fig-results2}), ORB feature~\cite{rublee2011orb} obviously outperforms the baseline and other features.
Third, with respect to \emph{efficiency} (see \tabref{tab-speed}), four binary features (ORB~\cite{rublee2011orb}, AKAZE~\cite{alcantarilla2011fast}, BRISK~\cite{leutenegger2011brisk}, FREAK~\cite{alahi2012freak}) outperform the baseline (SIFT~\cite{lowe2004distinctive}).

\textbf{d) matching solutions.}
As before, We regard SIFT matcher~\cite{lowe2004distinctive} to be the baseline.
First, with regard to \emph{matching ability} (see \tabref{tab-auc}), three rich feature matchers (CODE~\cite{lin2017code}, RepMatch~\cite{lin2016repmatch}, and GMS~\cite{bian2017gms}) outperform the baseline consistently.
KVLD matcher~\cite{liu2012virtual} beats the baseline in sequence 05 and is beaten by the latter in other sequences.
GAIM matcher~\cite{collins2014analysis} shows consistently lower performances than the baseline.
Second, with regard to \emph{correspondence sufficiency} (see \figref{fig-results1} or \figref{fig-results2}), three rich feature matchers (CODE~\cite{lin2017code}, RepMatch~\cite{lin2016repmatch}, and GMS~\cite{bian2017gms}) outperform the baseline, and other matchers (KVLD and GAIM) show similar performances with the baseline.
Third, with respect to \emph{efficiency} (see \tabref{tab-speed}), only GMS matcher shows higher speed than the baseline by adopting GPU acceleration, and other matchers are much more slower than the baseline.

\textbf{e) The best generic matcher}.
GMS matcher~\cite{bian2017gms} outperforms sparse matchers in terms of \emph{matching ability} and \emph{correspondence sufficiency}, although is weaker than other two rich matchers (CODE~\cite{lin2017code}, RepMatch~\cite{lin2016repmatch}).
With respect to \emph{efficiency}, it is several orders of magnitude faster than rich feature matchers (CODE and RepMatch), and is efficient enough to enable real-time performances by using GPU.
Therefore, we get the conclusion that GMS matcher~\cite{bian2017gms} shows \textbf{the best trade-off} among \emph{matching ability}, \emph{correspondence sufficiency}, and \emph{efficiency}.

\begin{table}
\centering
\caption{The time consumption of different matchers. Values in brackets mean GPU time, and others stand for CPU time.}
\resizebox{0.96\textwidth}{!}{%
\begin{tabular}{|c|c|c|c|c|}
\hline
Matchers & Feature numbers & Detection time (ms) & Matching time (ms)  & Selection time (ms)\\
\hline
SIFT & 1082 & 56.4 & 21.2 &  \multirow{11}{*}{1.0}\\
\cline{0-3}
SURF & 1432 & 63.0 & 19.7 & \\
\cline{0-3}
ORB & 3539 & 10.3 & 38.2 & \\
\cline{0-3}
AKAZE & 726 & 29.7 & 3.6 & \\
\cline{0-3}
BRISK & 1160 & 17.8 & 5.4 &  \\
\cline{0-3}
KAZE & 1060 & 187.0 & 15.5 &  \\
\cline{0-3}
DLCO & 1082 & 430.4 & 21.7 & \\
\cline{0-3}
FREAK & 919 & 43.9 & 3.2 & \\
\cline{0-3}
BinBoost & 1082 & 94.0 & 4.2 & \\
\cline{0-3}
LATCH & 984 & 86.1 & 3.3 & \\
\cline{0-3}
DAISY & 1082 & 79.3 & 25.2 & \\
\hline
KVLD & 1082 & 56.4 & 21.2 & 540.1 \\
\hline
GAIM & 45345 & 6783.2 & 1550.6 & 7145.1 \\
\hline
CODE & \multirow{2}{*}{64609} & \multirow{2}{*}{(1365.0)} & \multirow{2}{*}{(970.1)} & 3079.6 \\
\cline{0-0}\cline{5-5}
RepMatch & & & & 10779.6\\
\hline
GMS & 9463 & 33.0 & (12.4) & 1.3 \\
\hline
\end{tabular} %
}
\label{tab-speed}
\end{table}

\section{Discussion}\label{discussion}
Our primary goal is to set up an uniform benchmark to evaluate feature matchers.
We have made significant efforts on making it reasonable and convenient to use as well as possible.
The proposed benchmark is discussed below.

\textbf{Contribution and novelty}.
As introduced in \secref{intro}, the \textbf{contribution} of this paper includes 
i) we set up \textbf{the first uniform feature matching benchmark} to facilitate the evaluation of feature matchers, which enables researchers explore and develop their matchers conveniently. 
ii) we conduct exhaustive evaluation of different state-of-the-art matchers, where the results and conclusions can be used to design practical matchers in real applications and also advocate the potential future research directions in the filed of local feature extraction and matching solutions.
On the other hand, the \textbf{novelty} involves proposing three different aspects to evaluate matchers, designing corresponding evaluation metrics, and creating (re-organizing) benchmark datasets for enabling both \emph{short-baseline} and \emph{wide-baseline} feature matching evaluation. 

\textbf{Evaluation metrics}.
The proposed \textbf{SP curves} (with \textbf{AUC score}) and \textbf{AP bars} rely on \textbf{camera pose estimation} which we use to judge whether a pair is matched correctly.
Therefore, the performance of matching is not only leaded by feature matchers but also \textbf{pose estimators}.
One may concern that \emph{pose estimators} could not work perfectly and it would lead to an incorrect comparison of matchers.
For example, estimators may sometimes fail to get a correct pose estimation even though an image pair is matched well.
However, we argue that \textbf{the current solution is reasonable} because \emph{two-view pose estimation} is an essential part in SfM/Monocular SLAM where the estimated camera pose is directly used to initialize the system even though other pairs' poses could be refined in further processing when the system has been initialized.
Therefore, our current evaluation implies \textbf{how likely a matcher can enable correct initialization} in SfM or Monocular SLAM.
This is a very practical and vital problem! 

\textbf{Benchmark datasets}.
Although the benchmark dataset covers a wide range of scenes, one may concern that the images in \emph{wide-baseline portion} are not as diverse as images in some SfM datasets, like Internet-image collections~\cite{agarwal2011building} where images are captured from \textbf{many different cameras}.
We exclude these diverse datasets because they often cannot provide precise ground-truth camera positions for evaluation.
One possible solution to use these dataset for evaluation is \textbf{reconstructing 3D models using SfM tools} and regarding their estimated camera positions as "ground-truth".
However, we argue that \textbf{it is not reliable enough} and instead propose to \textbf{sub-sample video sequences with precise camera positions} for our evaluation.
Besides, even though the current single-camera setting is not as diverse as Internet-image datasets, \textbf{it is still practical in many real-life scenarios}.
For example, we sometimes may need to reconstruct 3D models of an office (or a living room) from unordered photos captured by a smart phone. 
Finally, we will still be considering how to introduce more diversified datasets while keeping the ground truth accurate.

\textbf{Evaluated methods}.
The proposed benchmark not only can be used to benchmark feature matchers but also \textbf{pose estimators}.
Since currently we are more interested in feature matchers, various pose estimators are not introduced in our evaluation .
In order to maximize matchers' performances, we have adopted two state-of-the-art pose estimators and \textbf{select the properest possible one} for each matcher.
Limited to the page length, we would explore more pose estimators and add more ablation studies in the future work.

\section{Conclusions}\label{conclusion}

This paper proposes the first uniform benchmark to evaluate feature matchers.
It suggests analyzing matchers in three different aspects, including matching ability, correspondence sufficiency, and efficiency.
In order to measure these different properties, the paper presents two novel evaluation metrics.
On the other hand, the proposed benchmark dataset covers a wide range of scenes and can be used to evaluate matchers in different type of problems, involving short-baseline matching and wide-baseline matching.
What's more, comprehensive evaluation of different feature matchers is carried out and results are useful for researchers to design practical matching systems in real applications.

\bibliographystyle{splncs}
\bibliography{egbib}

\begin{thebibliography}{10}

\bibitem{schonberger2016structure}
Schonberger, J.L., Frahm, J.M.:
\newblock Structure-from-motion revisited.
\newblock In: {IEEE Conference on Computer Vision and Pattern Recognition
  (CVPR)}. (2016)  4104--4113

\bibitem{davison2007monoslam}
Davison, A.J., Reid, I.D., Molton, N.D., Stasse, O.:
\newblock Monoslam: Real-time single camera slam.
\newblock {IEEE Transactions on Pattern Recognition and Machine Intelligence
  (PAMI)} \textbf{29} (2007)  1052--1067

\bibitem{mur2015orb}
Mur-Artal, R., Montiel, J.M.M., Tardos, J.D.:
\newblock {ORB-SLAM}: a versatile and accurate monocular slam system.
\newblock {IEEE Transactions on Robotics (TOR)} \textbf{31} (2015)  1147--1163

\bibitem{mikolajczyk2005comparison}
Mikolajczyk, K., Tuytelaars, T., Schmid, C., Zisserman, A., Matas, J.,
  Schaffalitzky, F., Kadir, T., Van~Gool, L.:
\newblock A comparison of affine region detectors.
\newblock {International Journal on Computer Vision (IJCV)} \textbf{65} (2005)
  43--72

\bibitem{moreels2007evaluation}
Moreels, P., Perona, P.:
\newblock Evaluation of features detectors and descriptors based on 3d objects.
\newblock {International Journal on Computer Vision (IJCV)} \textbf{73} (2007)
  263--284

\bibitem{mikolajczyk2005performance}
Mikolajczyk, K., Schmid, C.:
\newblock A performance evaluation of local descriptors.
\newblock {IEEE Transactions on Pattern Recognition and Machine Intelligence
  (PAMI)} \textbf{27} (2005)  1615--1630

\bibitem{heinly2012comparative}
Heinly, J., Dunn, E., Frahm, J.M.:
\newblock Comparative evaluation of binary features.
\newblock In: {European Conference on Computer Vision (ECCV)}.
\newblock Springer (2012)  759--773

\bibitem{hpatches_2017_cvpr}
Balntas, V., Lenc, K., Vedaldi, A., Mikolajczyk, K.:
\newblock {HPatches}: A benchmark and evaluation of handcrafted and learned
  local descriptors.
\newblock In: {IEEE Conference on Computer Vision and Pattern Recognition
  (CVPR)}, IEEE (2017)  5173--5182

\bibitem{schonberger2017comparative}
Sch{\"o}nberger, J.L., Hardmeier, H., Sattler, T., Pollefeys, M.:
\newblock Comparative evaluation of hand-crafted and learned local features.
\newblock In: {IEEE Conference on Computer Vision and Pattern Recognition
  (CVPR)}, IEEE (2017)  6959--6968

\bibitem{Geiger2012CVPR}
Geiger, A., Lenz, P., Urtasun, R.:
\newblock Are we ready for autonomous driving? the kitti vision benchmark
  suite.
\newblock In: {IEEE Conference on Computer Vision and Pattern Recognition
  (CVPR)}, IEEE (2012)  3354--3361

\bibitem{sturm12iros}
Sturm, J., Engelhard, N., Endres, F., Burgard, W., Cremers, D.:
\newblock A benchmark for the evaluation of rgb-d slam systems.
\newblock In: {IEEE International Conference on Intelligent Robots and Systems
  (IROS)}. (2012)

\bibitem{strecha2008benchmarking}
Strecha, C., Von~Hansen, W., Van~Gool, L., Fua, P., Thoennessen, U.:
\newblock On benchmarking camera calibration and multi-view stereo for high
  resolution imagery.
\newblock In: {IEEE Conference on Computer Vision and Pattern Recognition
  (CVPR)}, IEEE (2008)  1--8

\bibitem{lowe2004distinctive}
Lowe, D.G.:
\newblock Distinctive image features from scale-invariant keypoints.
\newblock {International Journal on Computer Vision (IJCV)} \textbf{60} (2004)
  91--110

\bibitem{bay2008speeded}
Bay, H., Ess, A., Tuytelaars, T., Van~Gool, L.:
\newblock Speeded-up robust features ({SURF}).
\newblock {Computer Vision and Image Understanding (CVIU)} \textbf{110} (2008)
  346--359

\bibitem{rublee2011orb}
Rublee, E., Rabaud, V., Konolige, K., Bradski, G.:
\newblock Orb: An efficient alternative to sift or surf.
\newblock In: {IEEE International Conference on Computer Vision (ICCV)}, IEEE
  (2011)  2564--2571

\bibitem{liu2012virtual}
Liu, Z., Marlet, R.:
\newblock Virtual line descriptor and semi-local matching method for reliable
  feature correspondence.
\newblock In: {British Machine Vision Conference (BMVC)}. (2012)  16--1

\bibitem{collins2014analysis}
Collins, T., Mesejo, P., Bartoli, A.:
\newblock An analysis of errors in graph-based keypoint matching and proposed
  solutions.
\newblock In: {European Conference on Computer Vision (ECCV)}, Springer (2014)
  138--153

\bibitem{lin2017code}
Lin, W.Y., Wang, F., Cheng, M.M., Yeung, S.K., Torr, P.H., Do, M.N., Lu, J.:
\newblock Code: Coherence based decision boundaries for feature correspondence.
\newblock {IEEE Transactions on Pattern Recognition and Machine Intelligence
  (PAMI)} (2017)

\bibitem{lin2016repmatch}
Lin, W.Y., Liu, S., Jiang, N., Do, M.N., Tan, P., Lu, J.:
\newblock Repmatch: Robust feature matching and pose for reconstructing modern
  cities.
\newblock In: {European Conference on Computer Vision (ECCV)}, Springer (2016)
  562--579

\bibitem{bian2017gms}
Bian, J., Lin, W.Y., Matsushita, Y., Yeung, S.K., Nguyen, T.D., Cheng, M.M.:
\newblock {GMS}: Grid-based motion statistics for fast, ultra-robust feature
  correspondence.
\newblock In: {IEEE Conference on Computer Vision and Pattern Recognition
  (CVPR)}, IEEE (2017)  4181--4190

\bibitem{schonberger2016vote}
Sch{\"o}nberger, J.L., Price, T., Sattler, T., Frahm, J.M., Pollefeys, M.:
\newblock A vote-and-verify strategy for fast spatial verification in image
  retrieval.
\newblock In: Asian Conference on Computer Vision, Springer (2016)  321--337

\bibitem{muja2009fast}
Muja, M., Lowe, D.G.:
\newblock Fast approximate nearest neighbors with automatic algorithm
  configuration.
\newblock VISAPP (1) \textbf{2} (2009) ~2

\bibitem{fischler1981random}
Fischler, M.A., Bolles, R.C.:
\newblock Random sample consensus: a paradigm for model fitting with
  applications to image analysis and automated cartography.
\newblock Communications of the ACM \textbf{24} (1981)  381--395

\bibitem{chum2005matching}
Chum, O., Matas, J.:
\newblock Matching with prosac-progressive sample consensus.
\newblock In: {IEEE Conference on Computer Vision and Pattern Recognition
  (CVPR)}. Volume~1., IEEE (2005)  220--226

\bibitem{torr2000mlesac}
Torr, P.H., Zisserman, A.:
\newblock Mlesac: A new robust estimator with application to estimating image
  geometry.
\newblock {Computer Vision and Image Understanding (CVIU)} \textbf{78} (2000)
  138--156

\bibitem{rousseeuw2005robust}
Rousseeuw, P.J., Leroy, A.M.:
\newblock Robust regression and outlier detection. Volume 589.
\newblock John wiley \& sons (2005)

\bibitem{raguram2013usac}
Raguram, R., Chum, O., Pollefeys, M., Matas, J., Frahm, J.M.:
\newblock Usac: a universal framework for random sample consensus.
\newblock {IEEE Transactions on Pattern Recognition and Machine Intelligence
  (PAMI)} \textbf{35} (2013)  2022--2038

\bibitem{nister2004efficient}
Nist{\'e}r, D.:
\newblock An efficient solution to the five-point relative pose problem.
\newblock {IEEE Transactions on Pattern Recognition and Machine Intelligence
  (PAMI)} \textbf{26} (2004)  756--770

\bibitem{alcantarilla2012kaze}
Alcantarilla, P.F., Bartoli, A., Davison, A.J.:
\newblock Kaze features.
\newblock In: {European Conference on Computer Vision (ECCV)}, Springer (2012)
  214--227

\bibitem{alcantarilla2011fast}
Alcantarilla, P.F., Solutions, T.:
\newblock Fast explicit diffusion for accelerated features in nonlinear scale
  spaces.
\newblock {IEEE Transactions on Pattern Recognition and Machine Intelligence
  (PAMI)} \textbf{34} (2011)  1281--1298

\bibitem{leutenegger2011brisk}
Leutenegger, S., Chli, M., Siegwart, R.Y.:
\newblock Brisk: Binary robust invariant scalable keypoints.
\newblock In: {IEEE International Conference on Computer Vision (ICCV)}, IEEE
  (2011)  2548--2555

\bibitem{alahi2012freak}
Alahi, A., Ortiz, R., Vandergheynst, P.:
\newblock Freak: Fast retina keypoint.
\newblock In: {IEEE Conference on Computer Vision and Pattern Recognition
  (CVPR)}, IEEE (2012)  510--517

\bibitem{simonyan2014learning}
Simonyan, K., Vedaldi, A., Zisserman, A.:
\newblock Learning local feature descriptors using convex optimisation.
\newblock {IEEE Transactions on Pattern Recognition and Machine Intelligence
  (PAMI)} \textbf{36} (2014)  1573--1585

\bibitem{leordeanu2005spectral}
Leordeanu, M., Hebert, M.:
\newblock A spectral technique for correspondence problems using pairwise
  constraints.
\newblock In: {IEEE International Conference on Computer Vision (ICCV)}.
  Volume~2., IEEE (2005)  1482--1489

\bibitem{zhou2013deformable}
Zhou, F., De~la Torre, F.:
\newblock Deformable graph matching.
\newblock In: {IEEE Conference on Computer Vision and Pattern Recognition
  (CVPR)}, IEEE (2013)  2922--2929

\bibitem{zhou2012factorized}
Zhou, F., De~la Torre, F.:
\newblock Factorized graph matching.
\newblock In: {IEEE Conference on Computer Vision and Pattern Recognition
  (CVPR)}, IEEE (2012)  127--134

\bibitem{morel2009asift}
Morel, J.M., Yu, G.:
\newblock Asift: A new framework for fully affine invariant image comparison.
\newblock SIAM Journal on Imaging Sciences \textbf{2} (2009)  438--469

\bibitem{hartley2003multiple}
Hartley, R., Zisserman, A.:
\newblock Multiple view geometry in computer vision.
\newblock Cambridge university press (2003)

\bibitem{trzcinski2013boosting}
Trzcinski, T., Christoudias, M., Fua, P., Lepetit, V.:
\newblock Boosting binary keypoint descriptors.
\newblock In: {IEEE Conference on Computer Vision and Pattern Recognition
  (CVPR)}, IEEE (2013)  2874--2881

\bibitem{levi2016latch}
Levi, G., Hassner, T.:
\newblock Latch: learned arrangements of three patch codes.
\newblock In: Applications of Computer Vision (WACV), IEEE (2016)  1--9

\bibitem{tola2010daisy}
Tola, E., Lepetit, V., Fua, P.:
\newblock Daisy: An efficient dense descriptor applied to wide-baseline stereo.
\newblock {IEEE Transactions on Pattern Recognition and Machine Intelligence
  (PAMI)} \textbf{32} (2010)  815--830

\bibitem{agarwal2011building}
Agarwal, S., Furukawa, Y., Snavely, N., Simon, I., Curless, B., Seitz, S.M.,
  Szeliski, R.:
\newblock Building rome in a day.
\newblock Communications of the ACM \textbf{54} (2011)  105--112

\end{thebibliography}

\end{document}